\documentclass[conference]{IEEEtran}
\IEEEoverridecommandlockouts
\usepackage{cite}
\usepackage{amsmath,amssymb,amsfonts}
\usepackage{algorithmic}
\usepackage{algorithm}
\newtheorem{proposition}{Proposition}
\usepackage{graphicx}
\usepackage{placeins}
\usepackage{textcomp}
\usepackage{array,booktabs,tabularx}
\usepackage[table]{xcolor}
\definecolor{EviEquationNote}{gray}{0.25}
\newcommand{\eviterm}[2]{%
  \underbrace{#1}_{\text{\normalfont\footnotesize\color{EviEquationNote}#2}}%
}
\definecolor{EviHighlight}{RGB}{239,244,248}
\newcommand{\evihighlight}{\rowcolor{EviHighlight}}
\newcommand{\evitable}{%
  \normalfont\rmfamily\fontsize{8}{9.6}\selectfont
  \setlength{\tabcolsep}{2.8pt}%
  \renewcommand{\arraystretch}{1.12}%
  \setlength{\heavyrulewidth}{0.6pt}%
  \setlength{\lightrulewidth}{0.4pt}%
  \setlength{\cmidrulewidth}{0.3pt}%
}
\newcommand{\evitablenote}{\normalfont\rmfamily\fontsize{7}{8.4}\selectfont}
\newcommand{\evimeansd}[2]{%
  \begingroup
  \renewcommand{\arraystretch}{0.9}%
  \begin{tabular}[t]{@{}r@{}}#1\\
    {\evitablenote\ensuremath{\pm}\,#2}%
  \end{tabular}%
  \endgroup
}
\newcolumntype{L}[1]{>{\raggedright\arraybackslash}p{#1}}
\newcolumntype{R}[1]{>{\raggedleft\arraybackslash}p{#1}}
\newcolumntype{G}[1]{>{\hspace*{\dimexpr#1-2\tabcolsep\relax}}r}
\newsavebox{\evitablebox}

\usepackage[hidelinks]{hyperref}
\hypersetup{pdftitle={Clinical Evidence-Guided Residual Learning for Medication Recommendation with Dynamic DDI Regularization},
pdfauthor={Zesheng Li, Chengchang Pan, Honggang Qi},
pdfkeywords={EviDrug, medication recommendation, clinical evidence, residual learning, dynamic DDI regularization}}
\begin{document}
\title{Clinical Evidence-Guided Residual Learning for Medication Recommendation with Dynamic DDI Regularization}
\author{\IEEEauthorblockN{Zesheng Li, Chengchang Pan, Honggang Qi}
\IEEEauthorblockA{School of Computer Science and Technology, University of Chinese Academy of Sciences, China\\
lizesheng26@mails.ucas.ac.cn,\quad 166353314@qq.com,\quad hgqi@ucas.ac.cn}}
\maketitle

\begin{abstract}
Medication recommendation must combine patient-specific treatment evidence with the interaction structure of a prescription. The central challenge is to determine which evidence supports each medication and how strongly to regularize interactions while preserving predictive quality. We propose EviDrug, a clinical evidence-guided residual learning framework. Its clinical-memory backbone integrates a self-trained visit-aware encoder, training-case memory and one medication-set interaction block to establish a patient-specific prescription. A gated refinement head combines the backbone states with clinical associations and history-conditioned medication identities through bounded logit corrections. Drug--drug interaction (DDI) feedback adapts regularization strength using training-query predictions. Across MIMIC-IV, NWICU and SICdb, EviDrug leads twelve published-method baselines in AP, F1 and Jaccard, improving AP by 2.26, 2.06 and 1.71 percentage points over the strongest comparator. These gains come at lower parameter and runtime costs than MetaDrug. Matched controls demonstrate DDI reductions with stable predictive quality.

Code is available at \url{https://github.com/LIZESHENG13/EviDrug}.
\end{abstract}

\section{Introduction}
Medication recommendation turns a patient's electronic health record (EHR) into a medication combination. Its supporting evidence spans current clinical codes, earlier prescriptions and other patients' encounters. Individual records describe particular patients and times, whereas population statistics summarize many encounters. The data engineering challenge is to \textbf{integrate evidence at these different granularities for each medication}, while accounting for interactions within the resulting prescription.

\begin{figure}[!t]
\centering
\includegraphics[width=\columnwidth]{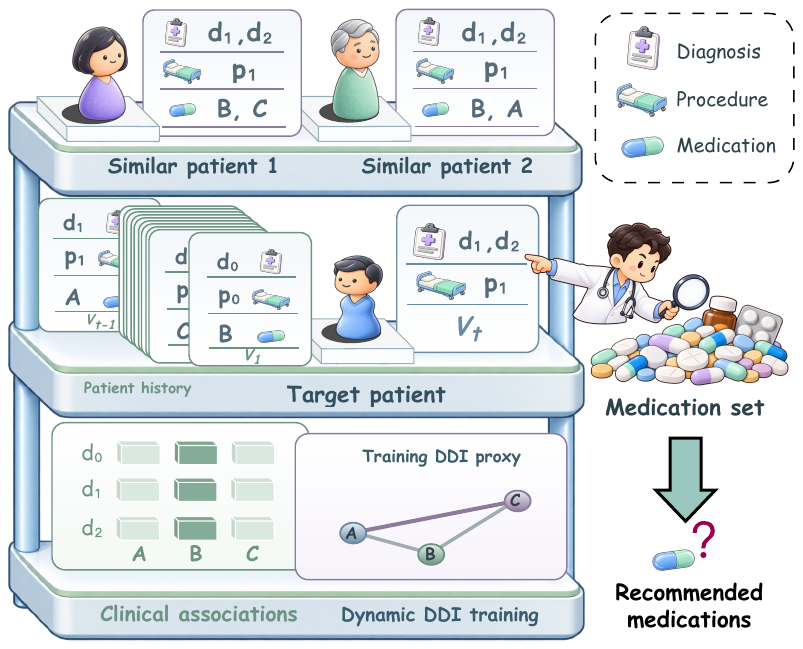}
\caption{Motivating example of medication-specific evidence and interaction control.}
\label{fig:overview}
\end{figure}

Existing methods use longitudinal memory and medication knowledge~\cite{shang2019gamenet}, history-aware set prediction~\cite{zhang2025sspnet}, and adaptation to a patient's own and similar patients' records~\cite{moghaddam2026metadrug}. These approaches establish the value of reusing clinical records. Yet \textbf{case relevance and medication support are different quantities}: a relevant encounter can contain medications with different levels of support for the target patient. Integrating its prescription requires candidate-level judgments, especially when personal history, neighboring cases and population associations favor different medications. The question is how to preserve these distinct sources of evidence while combining them into one recommendation.

Figure~\ref{fig:overview} illustrates this mismatch. The target encounter contains diagnoses $d_1,d_2$ and procedure $p_1$. Two clinically similar patients received $\{B,C\}$ and $\{B,A\}$, respectively, while the target's history already includes $B$. Both cases support $B$, but their agreement on the clinical context leaves different evidence for $A$ and $C$. Transferring an entire neighboring prescription can import weakly supported medications; relying on history alone can miss candidates supported by the current condition. Diagnosis--medication associations provide an additional population-level reference for these candidate-specific judgments. The DDI relations among $A$, $B$ and $C$ further show why improving individual scores must be coupled with control of the resulting combination.

Evidence integration and interaction control are consequently linked. DDI-aware models already combine prediction with interaction objectives, including adaptive loss balancing~\cite{yang2021safedrug,yang2023molerec}. Evidence refinement changes which medications are selected together, creating a need to adapt DDI regularization to the resulting combinations. The objective is to reduce the interaction rate while retaining the predictive evidence learned from patient trajectories and training cases.

EviDrug organizes the evidence into a training-case memory, clinical association statistics and the target patient's history. A clinical-memory backbone establishes a prescription from the history and retrieved encounters. A shared residual head then \textbf{aligns the evidence with individual medications} through bounded score corrections. Training cases and statistics form reusable references; each request reads them with fixed model parameters. DDI-feedback optimization controls interactions during training. This design connects evidence integration with recommendation quality and processing cost.

This paper makes three contributions:
\begin{itemize}
\item \textbf{Medication-wise evidence integration.} We organize training encounters and code--medication statistics as complementary references. Medication-specific case attention and a shared residual head integrate their evidence with patient history, making bounded corrections to the initial prescription.
\item \textbf{DDI feedback from predicted combinations.} We adjust the interaction penalty using the DDI rate of thresholded training predictions. The prediction-loss weights remain fixed, and feedback adds no parameters or forward passes at inference.
\item \textbf{Recommendation gains across three cohorts.} EviDrug improves AP over the strongest baseline by \textbf{2.26, 2.06 and 1.71 percentage points} on the full MIMIC-IV, NWICU and SICdb cohorts, respectively, while using fewer parameters and less inference time than MetaDrug.

\end{itemize}

\section{Related Work}
\label{sec:related-work}

\subsection{Clinical Representations}
Doctor AI and DeepCare use recurrent patient states for longitudinal prediction, with DeepCare additionally modeling irregular timing and intervention effects~\cite{choi2015doctorai,pham2016deepcare}. RETAIN exposes the contribution of visits and clinical variables through reverse-time attention, whereas Deepr uses convolution to extract local clinical patterns~\cite{choi2016retain,nguyen2017deepr}. Code representations can also draw on information beyond the observed sequence. GRAM shares information through a code's medical-ontology ancestors, allowing related concepts to borrow statistical strength when their own observations are sparse~\cite{choi2017gram}. BEHRT contextualizes diagnosis sequences using age, segment and position information, and Med-BERT pretrains structured-code representations on large EHR corpora~\cite{li2020behrt,rasmy2021medbert}. G-BERT combines ontology-aware embeddings with single-visit pretraining before longitudinal fine-tuning for medication recommendation~\cite{shang2019gbert}. Pretraining supplies initial clinical representations; history aggregation determines how earlier encounters enter a query. Medication recommendation further requires a readout from those states to candidate scores. The same encounter can supply different evidence for different medications, which motivates medication-specific processing after clinical encoding.

\subsection{Patient History and Case Evidence}
Clinical evidence can be reused within a patient's trajectory or across patients. DMNC maintains separate memories for asynchronous clinical sequences~\cite{le2018dmnc}. MICRON models medication changes through recurrent residual updates, while COGNet combines generation with copying from medication history~\cite{yang2021micron,wu2022cognet}. ARCI associates medical intents with temporal patterns through contrastive learning~\cite{moghaddam2024arci}. MeLU learns preferences from support examples in general recommendation~\cite{lee2019melu}. MetaDrug, presented in the ICDE 2026 Industry and Application track, applies self-adaptation to patient history and peer-adaptation to similar visits, with uncertainty filtering of support evidence~\cite{moghaddam2026metadrug}. EviDrug builds on this use of self and peer records by separating evidence retrieval from its medication-wise readout. A fixed training-case bank supplies reference encounters; attention weights their prescriptions separately for each medication. Aggregate code--medication statistics provide a second reference alongside individual cases. Shared model parameters combine these references with patient history without fitting a new model for each query. This extends record reuse from patient-level adaptation to candidate-level evidence integration.

\subsection{Medication Combinations}
LEAP constructs medication combinations through content attention and recurrent decoding, modeling both code--medication and medication--medication dependencies~\cite{zhang2017leap}. GAMENet places EHR co-prescription and DDI graphs in medication memory~\cite{shang2019gamenet}. SSPNet combines clinical set representations, treatment history and medication knowledge through a permutation-consistent decoder~\cite{zhang2025sspnet}. Deep Sets characterizes permutation-invariant representations, and Set Transformer models interactions among set elements using attention~\cite{zaheer2017deepsets,lee2019settransformer}. For medication recommendation, these relationships have different meanings. Co-prescription records describe which medications occur together in training encounters; attention between medication states depends on the current query. A diagnosis--medication association instead describes support for a candidate under a clinical code. Keeping these roles explicit allows a refinement head to apply candidate-level clinical evidence to an existing set prediction, with the patient and set encoders held fixed.

\subsection{DDI Regularization}
LEAP includes clinical knowledge in its reinforcement-learning reward~\cite{zhang2017leap}. SafeDrug combines molecular representations with interaction control~\cite{yang2021safedrug}, while MoleRec learns patient-conditioned molecular substructures and anneals the DDI-loss weight~\cite{yang2023molerec}. Adaptive DDI weighting is already part of medication recommendation. The question for a feedback rule is which predictions it monitors and which parameters it updates. Work on nondifferentiable rate constraints distinguishes a hard decision rate from a smooth proxy used for gradient updates~\cite{cotter2019constraints}. After thresholding, scores can move without changing a selected set. A smooth interaction loss can consequently change while the reported hard DDI rate stays constant. Whether two candidates are selected together determines whether their interaction contributes to the hard rate. Feedback on selected sets monitors the interaction statistic used in evaluation. EviDrug uses the smooth rate to train its residual head with the backbone fixed, and the hard rate on training queries to update the DDI coefficient after each epoch. The prediction-loss weights stay fixed throughout.

\section{Method}
\label{sec:method}

\begin{figure*}[t]
\centering
\includegraphics[width=\textwidth]{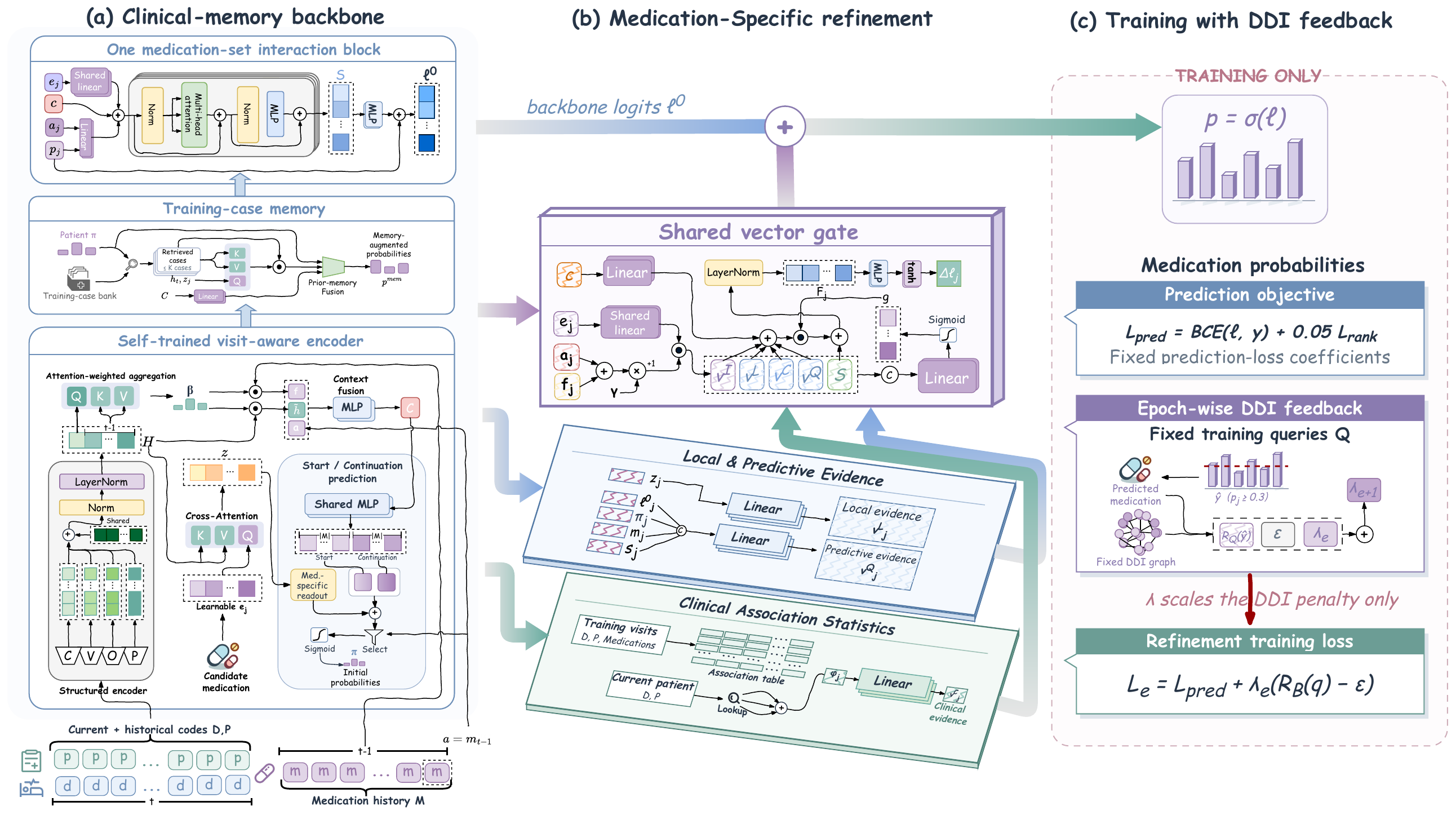}
\caption{EviDrug framework: (a) clinical-memory backbone, (b) medication-specific evidence refinement, and (c) DDI-feedback optimization during training.}
\label{fig:architecture}
\end{figure*}

EviDrug integrates evidence from patient histories, individual training encounters and population statistics. Figure~\ref{fig:architecture} follows the computation in three parts. The \textbf{clinical-memory backbone} retrieves training cases and combines their prescriptions with the patient's record. \textbf{Medication-specific refinement} aligns these backbone states with local clinical evidence and code--medication associations to correct each candidate's score. \textbf{DDI-feedback optimization} regulates interactions in the selected sets during training. At inference, fixed training references support one backbone-and-refinement path.

\subsection{Task Definition}
\label{sec:problem}
Let $\mathcal M$ contain $M$ medication classes. For a query encounter $T$, the available patient record is
\begin{equation}
\mathcal X=\left(\{\mathcal D_t,\mathcal P_t\}_{t=1}^{T},
\{\mathbf y_t\}_{t=1}^{T-1}\right),
\label{eq:task-input}
\end{equation}
where $\mathcal D_t$ and $\mathcal P_t$ are diagnosis and procedure codes, and $\mathbf y_t\in\{0,1\}^M$ records medication use. The medication-history matrix $\mathbf M_{<T}=[\mathbf y_1;\ldots;\mathbf y_{T-1}]$ supplies preceding treatments. EviDrug predicts the query medication set $\mathbf y_T$, which is used as supervision. Visit and medication-set states have width $d$; medication matching uses width $w$.

\subsection{Clinical-Memory Backbone}
\label{sec:clinical-memory-backbone}
Figure~\ref{fig:architecture}(a) combines three components: a self-trained visit-aware encoder, training-case memory and one medication-set interaction block. Their outputs are backbone logits $\boldsymbol\ell^0$ and medication states $\{\mathbf S_j\}_{j=1}^{M}$.

\subsubsection{Self-Trained Visit-Aware Encoder}
\label{sec:patient-prior}
The structured encoder learns \textbf{clinical-code and encounter-position embeddings} directly from the training cohort. Normalization and dropout form token states $\mathbf U_i$. With code positions $C_t$ and $n_t=|C_t|$, pooling and a visit MLP give
\begin{equation}
\mathbf h_t=f_v\!\left(\left[
P_h\frac{\sum_{i\in C_t}\mathbf U_i}{\sqrt{\max(1,n_t)}};
\log(1+n_t)\right]\right).
\label{eq:visit-encoding}
\end{equation}
The visit states form $\mathbf H=[\mathbf h_1;\ldots;\mathbf h_T]$. Query and retrieved encounters share this encoder; empty encounters receive zero states.

History attention uses the current encounter to select relevant preceding visits:
\begin{equation}
\begin{split}
\beta_t=\operatorname{softmax}_{t<T}\!\Bigl(
&\frac{(W_q^H\mathbf h_T)^\top W_k^H\mathbf h_t}{\sqrt d}\\
&{}-\eta_H\log(1+T-t)\Bigr),
\end{split}
\label{eq:history-attention}
\end{equation}
where $\eta_H=\operatorname{softplus}(r_H)>0$ learns the recency effect. The same weights summarize \textbf{prior clinical states and their recorded medications}:
\begin{align}
\widetilde{\mathbf h}&=\sum_{t<T}\beta_t\mathbf h_t,
\label{eq:history-clinical-summary}\\
\mathbf f&=\sum_{t<T}\beta_t\mathbf y_t.
\label{eq:history-medication-summary}
\end{align}
Let $\mathbf a=\mathbf y_{T-1}$ denote immediately preceding use. Context fusion combines current and preceding visit states, clinical change, $\widetilde{\mathbf h}$ and the treatment profile $(\mathbf a+\mathbf f)/2$ into $\mathbf c\in\mathbb R^d$. History-dependent inputs are zero when no preceding encounter is available.

Each candidate medication also reads current codes through cross-attention. Its learned identity $\mathbf e_j\in\mathbb R^w$ is the query; projected current code states $\widetilde{\mathbf U}_i=P_h\mathbf U_i$ supply keys and values:
\begin{align}
\omega_{ji}&=\operatorname{softmax}_{i\in C_T}\!\left(
\frac{\mathbf e_j^\top W_k^L\widetilde{\mathbf U}_i}{\sqrt w}\right),
\label{eq:local-attention}\\
\mathbf z_j&=\sum_{i\in C_T}\omega_{ji}
\tanh(W_v^L\widetilde{\mathbf U}_i).
\label{eq:local-readout}
\end{align}
The local state $\mathbf z_j$ captures the query codes supporting medication $j$ and is zero for an empty code set. A shared MLP maps $\mathbf c$ to start and continuation score vectors $[\mathbf b^{\mathrm{start}};\mathbf b^{\mathrm{cont}}]$. A medication-specific readout adds local evidence in each branch $u\in\{\mathrm{start},\mathrm{cont}\}$:
\begin{align}
\ell_j^{u}&=b_j^{u}+(\mathbf r_j^{u})^\top\mathbf z_j,
\label{eq:branch-logit}\\
\pi_j&=\sigma\!\left((1-a_j)\ell_j^{\mathrm{start}}+a_j\ell_j^{\mathrm{cont}}\right).
\label{eq:patient-prior}
\end{align}
The select operation in Figure~\ref{fig:architecture}(a) uses prior presence $a_j$ to choose the branch. Thus $\pi_j$ is a patient-specific probability before reading case memory; start means absent from the immediately preceding prescription.

\subsubsection{Training-Case Memory}
\label{sec:case-memory}
The case bank stores adjacent training-encounter transitions and their later medication indicators $\mathbf y_r^+$. Retrieval combines Jaccard rankings of the current clinical state and clinical changes into at most $K$ cases, $\mathcal R(\mathcal X)$. Cases from the query patient are excluded. Retrieval identifies related clinical situations; learned attention determines which prescriptions support each medication.

The shared encoder supplies donor states $\mathbf h_r^+$ and keys $\mathbf k_r^M=W_k^M\mathbf h_r^+$. A learned medication query $\mathbf q_j^M$ combines $\mathbf h_T$ and local evidence $\mathbf z_j$. Attention combines their match with current-code similarity $s_r^{\mathrm{code}}$:
\begin{equation}
\begin{split}
\beta_{rj}=\operatorname{softmax}_{r\in\mathcal R(\mathcal X)}\!\Bigl(
&\frac{(\mathbf q_j^M)^\top\mathbf k_r^M}{\sqrt w}\\
&{}+\operatorname{softplus}(r_M)s_r^{\mathrm{code}}\Bigr).
\end{split}
\label{eq:case-attention}
\end{equation}
Write $e_{rj}$ for the score inside this softmax. The values are recorded donor medications, giving $\overline y_j=\sum_r\beta_{rj}y_{rj}^+$. Prior-memory fusion uses strength $s_j=4\sigma(f_s(\mathbf c)_j)$:
\begin{equation}
p_j^{\mathrm{mem}}=
\frac{\eviterm{\pi_j}{patient prior}
+s_j\eviterm{\overline y_j}{case evidence}}{1+s_jm_j}.
\label{eq:memory-mixture}
\end{equation}
The attention mass $m_j=\sum_r\beta_{rj}$ is one for a nonempty shortlist and zero otherwise. Consequently, an empty memory returns the patient prior.

\begin{proposition}[Medication-specific memory supervision]
\label{prop:case-gradient}
For a fixed nonempty shortlist, let $G_j$ be the total downstream loss derivative with respect to the fusion probability in Eq.~\eqref{eq:memory-mixture}, before numerical clipping. Holding the other inputs fixed,
\begin{equation}
\frac{\partial\mathcal L}{\partial e_{rj}}
=G_j\frac{s_j}{1+s_j}\beta_{rj}
\bigl(y_{rj}^+-\overline y_j\bigr).
\label{eq:case-gradient}
\end{equation}
\end{proposition}
\begin{IEEEproof}
Softmax differentiation gives $\partial\overline y_j/\partial e_{rj}=\beta_{rj}(y_{rj}^+-\overline y_j)$. Since $m_j=1$, fusion has derivative $s_j/(1+s_j)$ with respect to $\overline y_j$. The chain rule yields Eq.~\eqref{eq:case-gradient}.
\end{IEEEproof}
When increasing a medication's probability reduces the downstream loss ($G_j<0$), the attention-score gradient favors cases supporting that medication above the donor average. The signal reverses when decreasing its probability is preferred. Thus a shared shortlist receives \textbf{medication-specific supervision} during backbone training.

\subsubsection{One Medication-Set Interaction Block}
The memory probabilities initialize one token per candidate, together with medication identity, prior use and patient context:
\begin{equation}
\mathbf t_j=P_e\mathbf e_j+P_s[p_j^{\mathrm{mem}};a_j]+\mathbf c,
\label{eq:medication-token}
\end{equation}
where $P_e$ and $P_s$ project medication identity and scalar evidence to width $d$. A \textbf{single four-head self-attention block} with a feed-forward sublayer maps these tokens to $\{\mathbf S_j\}_{j=1}^{M}$. Its readout adjusts the memory score:
\begin{equation}
\ell_j^0=\eviterm{\operatorname{logit}(p_j^{\mathrm{mem}})}{memory score}
+\eviterm{2\tanh(f_{\mathrm{set}}(\mathbf S_j))}{set interactions}.
\label{eq:set-backbone}
\end{equation}
The backbone parameters $\theta_b$ are learned jointly. The identity $\mathbf e_j$ and projection $P_e$ are reused by refinement, linking the evidence update to the same medication representation.

\subsection{Medication-Specific Refinement}
\label{sec:evidence-refinement}
Figure~\ref{fig:architecture}(b) refines the established prescription using complementary evidence for each medication. The \textbf{backbone is fixed} during this stage. Its local clinical states and prediction context form
\begin{align}
\mathbf v_j^L&=W_L\mathbf z_j,
\label{eq:local-evidence}\\
\mathbf v_j^Q&=P_Q[\ell_j^0;\pi_j;m_j;s_j].
\label{eq:predictive-evidence}
\end{align}
These are the local and predictive evidence branches in the figure. They tell the refinement head which current codes support a medication and how the backbone and memory have scored it.

Clinical association statistics provide a population-level reference from training visits. Let $N$ be the visit count; $n_j$, $n_c$ and $n_{jc}$ count visits containing medication $j$, code $c$, and both, respectively. The smoothed association is
\begin{equation}
L_{jc}=\log\frac{n_{jc}+b}{n_j+2b}
-\log\frac{n_c-n_{jc}+b}{N-n_j+2b}.
\label{eq:association}
\end{equation}
For each query code group, lookup and aggregation sum these associations for every medication. Standardizing the sums across medications gives relative support; dividing by the code count gives average support. These two summaries for all codes, diagnoses and procedures form $\boldsymbol\phi_j\in\mathbb R^6$, projected to $\mathbf v_j^C=W_C\boldsymbol\phi_j$. Training queries exclude their own patient's visits from the counts; validation and test queries read fixed training statistics.

History-conditioned medication identity supplies the patient's treatment context:
\begin{equation}
\mathbf v_j^I=\left(1+\gamma\frac{a_j+f_j}{2}\right)P_e\mathbf e_j.
\label{eq:personalized-identity}
\end{equation}
The learned scalar $\gamma$ modulates medication identity with immediately preceding and attention-weighted historical use. A shared vector gate then determines how the evidence updates the medication state:
\begin{align}
\mathbf v_j&=\mathbf v_j^L+\mathbf v_j^I+\mathbf v_j^Q
+\mathbf v_j^C+W_c\mathbf c,
\label{eq:evidence-aggregate}\\
\mathbf g_j&=\sigma\!\left(P_g[\mathbf S_j;\mathbf v_j^L;
\mathbf v_j^I;\mathbf v_j^Q;\mathbf v_j^C]\right).
\label{eq:evidence-gate}
\end{align}
All evidence states have width $d$. Gate parameters are \textbf{shared across medications}, while $\mathbf g_j$ depends on each medication's evidence. The gated state $\mathbf F_j=\operatorname{LN}(\mathbf S_j+\mathbf g_j\odot\mathbf v_j)$ produces
\begin{align}
\Delta\ell_j&=2\tanh(f_o(\mathbf F_j)),
\label{eq:bounded-correction}\\
\ell_j&=\eviterm{\ell_j^0}{backbone}
+\eviterm{\Delta\ell_j}{evidence correction}.
\label{eq:residual-update}
\end{align}
The output is $p_j=\sigma(\ell_j)$. Zero-initializing the final readout layer makes refinement \textbf{begin exactly at the backbone prediction}.

The bounded correction also controls its influence on the prescription. For backbone probability $p_j^0=\sigma(\ell_j^0)$ and odds $O(p)=p/(1-p)$,
\begin{equation}
e^{-2}\ \leq\ \frac{O(p_j)}{O(p_j^0)}\ \leq\ e^2.
\label{eq:odds-bound}
\end{equation}
Indeed, the odds ratio equals $\exp(\ell_j-\ell_j^0)=\exp(\Delta\ell_j)$, and $|\Delta\ell_j|\leq2$. At threshold $\tau$, a backbone decision with $|\ell_j^0-\operatorname{logit}(\tau)|>2$ is therefore unchanged by refinement. This establishes the score region in which evidence can revise a recommendation.

\subsection{DDI-Feedback Optimization}
\label{sec:dynamic-training}
Figure~\ref{fig:architecture}(c) separates prediction supervision from the feedback that adjusts DDI regularization. For a training minibatch $B$, prediction uses binary cross-entropy and a ranking term:
\begin{equation}
\mathcal L_{\mathrm{pred}}(\boldsymbol\ell)
=\mathcal L_{\mathrm{BCE}}(\boldsymbol\ell,\mathbf y)
+0.05\mathcal L_{\mathrm{rank}}(\boldsymbol\ell,\mathbf y),
\label{eq:prediction-loss}
\end{equation}
where $\mathcal L_{\mathrm{rank}}$ averages softplus penalties when an unobserved medication outranks an observed one. Its coefficient $0.05$ is fixed. Backbone learning uses $\mathcal L_{\mathrm{pred}}(\boldsymbol\ell^0)$ with auxiliary BCE on $\boldsymbol\pi$. With the backbone frozen, the refinement head first learns the prediction objective $\mathcal L_{\mathrm{pred}}(\boldsymbol\ell)$; DDI-feedback optimization then adds interaction regularization to the same head.

Let $D$ be the fixed symmetric DDI adjacency with zero diagonal. A differentiable selection weight approximates thresholding at $\tau$:
\begin{equation}
q_{ij}=\sigma\!\left(\frac{\ell_{ij}-\operatorname{logit}(\tau)}{T_s}\right).
\label{eq:soft-selection}
\end{equation}
The matrix $q$ contains soft selection weights for minibatch $B$, with $q_{ij}=1/2$ at threshold $\tau$ and transition temperature $T_s>0$. These weights approximate the hard selection used for recommendation. The pooled interaction rate is
\begin{equation}
R_B(q)=\frac{\sum_{i\in B}\sum_{j<k}D_{jk}q_{ij}q_{ik}}
{\max(1,\sum_{i\in B}\sum_{j<k}q_{ij}q_{ik})}.
\label{eq:pooled-ddi}
\end{equation}
The numerator is soft interacting-pair mass, and the denominator is soft mass over all unordered pairs. Their ratio regularizes interaction frequency in the selected combination.

\begin{proposition}[Interaction-rate gradient]
\label{prop:ddi-gradient}
Write $Z_B=\sum_{i\in B}\sum_{j<k}q_{ij}q_{ik}$. For $Z_B>1$, define $u_{ij}=\sum_{k\ne j}D_{jk}q_{ik}$ and $v_{ij}=\sum_{k\ne j}q_{ik}$. Then
\begin{equation}
\frac{\partial R_B}{\partial\ell_{ij}}
=\frac{q_{ij}(1-q_{ij})}{T_s Z_B}
\bigl(u_{ij}-R_B(q)v_{ij}\bigr).
\label{eq:ddi-gradient}
\end{equation}
\end{proposition}
\begin{IEEEproof}
Let $A_B$ be the numerator of Eq.~\eqref{eq:pooled-ddi}. Symmetry and the zero diagonal of $D$ give $\partial A_B/\partial q_{ij}=u_{ij}$ and $\partial Z_B/\partial q_{ij}=v_{ij}$. Differentiating $A_B/Z_B$ and using $\partial q_{ij}/\partial\ell_{ij}=q_{ij}(1-q_{ij})/T_s$ gives Eq.~\eqref{eq:ddi-gradient}.
\end{IEEEproof}
For positive DDI weight, gradient descent on the rate in logit space lowers scores whose weighted neighborhood rate $u_{ij}/v_{ij}$ exceeds the pooled rate and raises those below it. This distributes the DDI penalty according to the combination's structure. For $Z_B<1$, the denominator floor is constant and the derivative is $q_{ij}(1-q_{ij})u_{ij}/T_s$.

At epoch $e$ of DDI-feedback optimization, the minibatch objective is
\begin{equation}
\mathcal L_e=\eviterm{\mathcal L_{\mathrm{pred}}}{prediction}
+\eviterm{\lambda_e\bigl(R_B(q)-\epsilon\bigr)}{DDI regularization}.
\label{eq:dynamic-loss}
\end{equation}
The target $\epsilon$ is fixed; $\lambda_e\geq0$ is held constant within the epoch. \textbf{Only the DDI coefficient adapts}, leaving the prediction-loss coefficients fixed.

Epoch-wise feedback measures actual thresholded recommendations on a fixed set $\mathcal Q$ containing one sampled query prefix per training patient. It is a monitoring set of training queries, distinct from minibatch $B$. For $\widehat y_{ij}=\mathbf1[p_{ij}\geq\tau]$, the initial refinement model sets
\begin{equation}
\epsilon=0.9R_{\mathcal Q}(\widehat y^{\mathrm{init}}).
\label{eq:controller-target}
\end{equation}
Here $R_{\mathcal Q}$ applies the same pooled-pair rate to hard selections over $\mathcal Q$. After epoch $e$, the measured rate $r_e=R_{\mathcal Q}(\widehat y^{(e)})$ updates the next epoch's coefficient:
\begin{equation}
\lambda_{e+1}=\operatorname{clip}_{[0,\lambda_{\max}]}\!\bigl(
\lambda_e+\eta_\lambda\eviterm{(r_e-\epsilon)}{feedback error}\bigr).
\label{eq:coefficient-update}
\end{equation}
Rates above the target increase the DDI coefficient; rates below it decrease the coefficient. Smooth selections provide minibatch gradients, while fixed training queries provide epoch-level feedback on thresholded medication sets.

Algorithm~\ref{alg:dynamic-training} summarizes the learning procedure. Backbone and initial refinement models are selected by maximum validation AP. DDI-feedback optimization uses the initial refinement model as its reference and keeps the backbone fixed. The final model minimizes validation DDI subject to prespecified AP, F1, Jaccard and prescription-size bounds; the reference model remains a candidate. Section~\ref{sec:experiments} gives training budgets and selection tolerances; supplementary material specifies tie-breaking rules.

At inference, a clinical-code index retrieves reference encounters from the training-case bank. Their encoded keys and medication labels, together with the association table, are prepared once for the fixed model. Each query retrieves cases, aggregates code--medication statistics and produces $\{j:p_j\geq\tau\}$ through the shared backbone and refinement head. Medication-set attention costs $O(M^2d)$; dynamic feedback adds \textbf{zero inference parameters or forward passes}. The efficiency evaluation includes online retrieval, evidence preparation and prediction.

\begin{algorithm}[t]
\caption{EviDrug learning with DDI-feedback optimization}
\label{alg:dynamic-training}
\small
\textbf{Input:} training patients $\mathcal D_{\mathrm{tr}}$, validation patients $V$, DDI graph $D$, threshold $\tau$, stage budget $E$.\par
\textbf{Output:} backbone $\theta_b$ and refinement head $\theta_f^*$.\par\smallskip
\setlength{\tabcolsep}{2.4pt}
\renewcommand{\arraystretch}{1.05}
\begin{tabularx}{\linewidth}{@{}rX@{}}
\rowcolor{EviHighlight}[0pt][0pt]\multicolumn{2}{@{}l@{}}{\strut\textbf{Clinical-memory learning}}\\
1: & Build training-case memory and association statistics from $\mathcal D_{\mathrm{tr}}$; exclude query-patient cases and counts when reading evidence.\\
2: & Learn $\theta_b$ for $E$ epochs with $\mathcal L_{\mathrm{pred}}$ and auxiliary prior BCE.\\
3: & Select $\theta_b$ by validation AP; freeze it in evaluation mode.\\
\rowcolor{EviHighlight}[0pt][0pt]\multicolumn{2}{@{}l@{}}{\strut\textbf{Medication-specific refinement}}\\
4: & Initialize $\theta_f$ with a zero-valued final readout.\\
5: & Learn $\theta_f$ for $E$ epochs with $\mathcal L_{\mathrm{pred}}$.\\
6: & Select $\theta_f^{(0)}$ by validation AP; set $\mathcal C\gets\{\theta_f^{(0)}\}$.\\
\rowcolor{EviHighlight}[0pt][0pt]\multicolumn{2}{@{}l@{}}{\strut\textbf{DDI-feedback optimization}}\\
7: & Fix training queries $\mathcal Q$; set $\epsilon$ by Eq.~\eqref{eq:controller-target} and $\lambda_1\gets1$.\\
8: & \textbf{for} epoch $e=1,\ldots,E$ \textbf{do}\\
9: & \hspace*{1em}\textbf{for} each training minibatch $B$ \textbf{do}\\
10: & \hspace*{2em}Compute $q$ by Eq.~\eqref{eq:soft-selection}; update $\theta_f$ with $\mathcal L_e$ in Eq.~\eqref{eq:dynamic-loss}.\\
11: & \hspace*{1em}\textbf{end for}\\
12: & \hspace*{1em}Threshold predictions on $\mathcal Q$; measure $r_e\gets R_{\mathcal Q}(\widehat y^{(e)})$.\\
13: & \hspace*{1em}Compute $\lambda_{e+1}$ by Eq.~\eqref{eq:coefficient-update}.\\
14: & \hspace*{1em}Add $\theta_f^{(e)}$ to $\mathcal C$ if validation quality and prescription-size bounds hold.\\
15: & \textbf{end for}\\
16: & Select $\theta_f^*$ from $\mathcal C$ by minimum validation DDI, then maximum AP, then earliest epoch.\\
\end{tabularx}
\end{algorithm}

\section{Experimental Setup}
\label{sec:experiments}
\subsection{Cohorts and Patient Partitions}
We evaluate EviDrug on MIMIC-IV 3.1~\cite{physionetmimic31}, Northwestern ICU (NWICU) 0.1.0~\cite{physionetnwicu010} and SICdb 1.0.8~\cite{physionetsicdb108,pollardphysionet2026}. Table~\ref{tab:datasets} summarizes the complete eligible cohorts. MIMIC-IV contains 86,670 patients and 352,967 admissions; eligibility requires at least two chronologically valid admissions with diagnoses and mapped ATC3 medications.

\begin{table}[!t]
\centering
\caption{Cohort statistics and patient partitions.}
\label{tab:datasets}
\evitable
\begin{tabular*}{\columnwidth}{@{\extracolsep{\fill}}lrrrrr@{}}
\toprule
Dataset & Patients & Train/Val/Test & Codes & ATC3 & Map. (\%) \\
\midrule
MIMIC-IV & 86,670 & 62,402/6,934/17,334 & 35,391 & 201 & --- \\
NWICU & 12,492 & 8,993/1,000/2,499 & 6,258 & 167 & 75.03 \\
SICdb & 3,898 & 2,806/312/780 & 1,069 & 81 & 80.69 \\
\bottomrule
\end{tabular*}
\end{table}

Patients are partitioned into approximately 72\%/8\%/20\% train/validation/test, with the same partition shared by all methods within each cohort. Models train independently across cohorts. Vocabularies, association statistics and case memories use training patients only; clinical vocabulary counts include special tokens. The task reconstructs the medication set of each patient's final eligible encounter from its clinical codes and preceding history.

MIMIC-IV provides diagnoses and procedures; NWICU provides hospital ICD-10 diagnoses without a released procedure-ICD table; SICdb provides its main diagnosis without procedures. NWICU labels are prescription types with valid intervals overlapping the admission, whereas SICdb labels are positive-amount administered drug types. Unique ATC3 mapping retains 75.03\%/80.69\% of the corresponding medication records before cohort and vocabulary filtering. Complete eligibility, temporal-ordering and mapping rules are provided in the supplementary material.

The DDI graph uses the GAMENet interaction resource~\cite{shang2019gamenet}, derived from TWOSIDES~\cite{tatonetti2012twosides} and distributed by PyHealth~\cite{yang2023pyhealth}. We map its drug pairs to ATC3 classes, retain pairs within each cohort's training medication vocabulary, and construct a binary, symmetric adjacency matrix with zero diagonal. This fixed graph defines interaction regularization and DDI evaluation across methods.

\begin{table*}[!t]
\centering
\caption{Test performance on the full cohorts. Bold indicates the best value.}
\label{tab:published-baselines}
\evitable
\setlength{\tabcolsep}{2.5pt}
\begin{tabular}{llG{6pt}rrrG{8pt}rrrG{8pt}rrr}
\toprule
& & \multicolumn{4}{c}{MIMIC-IV} & \multicolumn{4}{c}{NWICU} & \multicolumn{4}{c}{SICdb} \\
\cmidrule(lr){3-6}\cmidrule(lr){7-10}\cmidrule(lr){11-14}
Method & Year / venue & AP & F1 & Jaccard & DDI & AP & F1 & Jaccard & DDI & AP & F1 & Jaccard & DDI \\
\midrule
RETAIN~\cite{choi2016retain} & 2016 / NeurIPS & 0.7034 & 0.5844 & 0.4330 & 0.0648 & 0.5840 & 0.5020 & 0.3503 & 0.1341 & 0.7064 & 0.6018 & 0.4522 & 0.1427 \\
Deepr~\cite{nguyen2017deepr} & 2017 / IEEE JBHI & 0.6917 & 0.5694 & 0.4168 & 0.0623 & 0.5659 & 0.4912 & 0.3402 & 0.1316 & 0.7053 & 0.5997 & 0.4498 & 0.1348 \\
Transformer~\cite{vaswani2017attention} & 2017 / NeurIPS & 0.6940 & 0.5652 & 0.4133 & 0.0679 & 0.5843 & 0.5000 & 0.3483 & 0.1256 & 0.7033 & 0.5972 & 0.4465 & 0.1312 \\
GAMENet~\cite{shang2019gamenet} & 2019 / AAAI & 0.6852 & 0.5715 & 0.4194 & 0.0684 & 0.5791 & 0.5037 & 0.3520 & 0.1216 & 0.7081 & 0.5919 & 0.4407 & 0.1356 \\
MeLU~\cite{lee2019melu} & 2019 / KDD & 0.6589 & 0.5358 & 0.3851 & 0.0644 & 0.5615 & 0.4738 & 0.3258 & 0.1392 & 0.7084 & 0.5874 & 0.4385 & 0.1433 \\
MICRON~\cite{yang2021micron} & 2021 / IJCAI & 0.7068 & 0.5853 & 0.4340 & 0.0631 & 0.5554 & 0.4817 & 0.3327 & 0.1499 & 0.7086 & 0.6055 & 0.4567 & 0.1398 \\
SafeDrug~\cite{yang2021safedrug} & 2021 / IJCAI & 0.6901 & 0.5613 & 0.4074 & 0.0567 & 0.5268 & 0.4574 & 0.3100 & \textbf{0.0802} & 0.6677 & 0.5480 & 0.3955 & 0.0583 \\
COGNet~\cite{wu2022cognet} & 2022 / WWW & 0.6083 & 0.5618 & 0.4248 & 0.0553 & 0.6030 & 0.5321 & 0.3862 & 0.1277 & 0.6950 & 0.6154 & 0.4693 & 0.1558 \\
MoleRec~\cite{yang2023molerec} & 2023 / WWW & 0.7000 & 0.5742 & 0.4199 & 0.0659 & 0.5610 & 0.4792 & 0.3288 & 0.0822 & 0.6576 & 0.5563 & 0.4052 & \textbf{0.0552} \\
ARCI~\cite{moghaddam2024arci} & 2024 / CIKM & 0.7006 & 0.5176 & 0.3652 & 0.0575 & 0.6268 & 0.5280 & 0.3756 & 0.0935 & 0.6920 & 0.6058 & 0.4553 & 0.1029 \\
SSPNet~\cite{zhang2025sspnet} & 2025 / IJCAI & 0.7536 & 0.6232 & 0.4725 & 0.0590 & 0.6703 & 0.5749 & 0.4213 & 0.0903 & 0.7334 & 0.6256 & 0.4774 & 0.1001 \\
MetaDrug~\cite{moghaddam2026metadrug} & 2026 / ICDE & 0.7082 & 0.5866 & 0.4370 & 0.0574 & 0.6186 & 0.5293 & 0.3766 & 0.1264 & 0.7198 & 0.6086 & 0.4602 & 0.1328 \\
\evihighlight
\textbf{EviDrug} & Ours & \evimeansd{\textbf{0.7763}}{0.0004} & \evimeansd{\textbf{0.6452}}{0.0022} & \evimeansd{\textbf{0.4988}}{0.0026} & \evimeansd{\textbf{0.0522}}{0.0014} & \evimeansd{\textbf{0.6909}}{0.0007} & \evimeansd{\textbf{0.6026}}{0.0016} & \evimeansd{\textbf{0.4515}}{0.0016} & \evimeansd{0.1043}{0.0032} & \evimeansd{\textbf{0.7505}}{0.0010} & \evimeansd{\textbf{0.6425}}{0.0023} & \evimeansd{\textbf{0.4950}}{0.0026} & \evimeansd{0.1181}{0.0052} \\
\bottomrule
\end{tabular}
\end{table*}

\subsection{Training, Controls, and Selection}
All EviDrug parameters are trained from scratch. Visit/set width is 128 and medication matching width is 32. Each of three optimization seeds uses 20 epochs of backbone learning, 20 of evidence refinement and 20 of DDI-feedback optimization. The backbone remains fixed throughout refinement and DDI optimization. Training uses AdamW~\cite{loshchilov2017adamw} with batch size 64. Regularization controls share the same refinement initialization within each seed.

Backbone and initial refinement models are selected by maximum validation AP. DDI optimization selects the lowest validation DDI while limiting AP decrease to 0.002, F1/Jaccard decrease to 0.005 each, and medication-count change to $\pm10\%$ relative to the initial refinement model. Dynamic feedback uses training-query predictions and targets 90\% of the initial training DDI. Full architecture, optimizer, feedback and tie-breaking settings are provided in the supplementary material.

We report patient-average AP, F1 and Jaccard, pooled unordered-pair DDI and mean medication count. Threshold 0.3 is fixed except for native COGNet decoding. EviDrug means and sample SDs summarize three optimization seeds on one split. Interacting-pair counts complement the DDI rate.

The twelve baselines use their documented single-seed training budgets and native inputs/decoding; EviDrug uses three seeds with the budget above.

\section{Experimental Results}
\label{sec:results}
The experiments evaluate evidence integration through prediction quality, complete recommendation cost, evidence-component ablations and DDI controls. Clinical analyses examine when patient history and population associations support continued and new-to-record medications.

\subsection{Medication Recommendation Across Cohorts}
Table~\ref{tab:published-baselines} tests whether medication-specific evidence refinement improves prediction across clinical cohorts. It compares EviDrug with twelve published methods on shared patient partitions. Complete MetaDrug includes self/peer adaptation and uncertainty filtering.

EviDrug leads all twelve baselines in \textbf{AP, F1 and Jaccard on every cohort}. Its AP gains over the strongest comparator, SSPNet, are \textbf{2.26, 2.06 and 1.71 percentage points} on MIMIC-IV, NWICU and SICdb. This consistent advantage spans ranking quality and thresholded prescription overlap. On MIMIC-IV, EviDrug combines these gains with the lowest DDI in the comparison. Across cohorts, DDI records the interaction profile at each method's operating point; the matched regularization controls in Table~\ref{tab:weights} isolate the effect of EviDrug's interaction objective.

The agreement among AP, F1 and Jaccard connects two aspects of the recommendation task. AP evaluates how observed medications are ranked among the candidates, whereas F1 and Jaccard evaluate the combination returned by the decision rule. EviDrug improves both the ordering of candidates and their overlap with the recorded prescription. In particular, its F1 gains over SSPNet are 2.20, 2.77 and 1.69 percentage points across the three cohorts; the corresponding Jaccard gains are 2.63, 3.02 and 1.76 points. Medication-specific evidence therefore improves the final set as well as its underlying scores.

The cohort comparison also tests the design under different clinical inputs. MIMIC-IV supplies both diagnoses and procedures, while NWICU and SICdb supply diagnosis information without procedure codes. EviDrug retains the same evidence-refinement design and learns each cohort independently. Its consistent ranking across these settings shows that the framework can combine available structured codes with treatment history, rather than requiring a particular procedure vocabulary. The component analyses below identify which evidence sources account for this advantage.

\begin{table}[!t]
\centering
\caption{Parameter counts (M) and inference time (ms/patient).}
\label{tab:efficiency}
\evitable
\begingroup
\newcommand{\evicostrows}{%
\toprule
 & \multicolumn{2}{c}{MIMIC-IV} & \multicolumn{2}{c}{NWICU} & \multicolumn{2}{c}{SICdb} \\
\cmidrule(lr){2-3}\cmidrule(lr){4-5}\cmidrule(lr){6-7}
Method & Params & Time & Params & Time & Params & Time \\
\midrule
RETAIN & 19.934 & 0.88 & 5.001 & 0.55 & 2.300 & 0.33 \\
Deepr & 18.617 & 1.06 & 3.683 & 0.41 & 0.983 & 0.15 \\
Transformer & 19.803 & 0.69 & 4.869 & 0.40 & 2.168 & 0.22 \\
GAMENet & 4.780 & 0.29 & 1.043 & 0.20 & 0.356 & 0.14 \\
MeLU & 2.289 & 0.57 & 0.423 & 0.37 & 0.087 & 0.23 \\
MICRON & 18.369 & 0.34 & 3.444 & 0.16 & 0.765 & 0.08 \\
SafeDrug & 5.858 & 0.27 & 1.992 & 0.20 & 0.900 & 0.10 \\
COGNet & 5.588 & 89.70 & 1.853 & 20.13 & 1.172 & 5.44 \\
MoleRec & 4.971 & 0.80 & 1.236 & 0.65 & 0.530 & 0.28 \\
ARCI & 1.181 & 17.19 & 1.163 & 12.08 & 1.119 & 4.20 \\
SSPNet & 5.106 & 6.86 & 1.373 & 3.67 & 0.698 & 0.97 \\
MetaDrug & 11.730 & 22.43 & 4.254 & 6.84 & 2.882 & 2.05 \\
\evihighlight
EviDrug & 7.766 & 10.84 & 2.142 & 3.25 & 1.069 & 1.38 \\
\bottomrule}
\sbox{\evitablebox}{\begin{tabular}{l*{3}{rr}}\evicostrows\end{tabular}}
\addtolength{\tabcolsep}{\dimexpr(\columnwidth-\wd\evitablebox)/14\relax}
\begin{tabular}{l*{3}{rr}}\evicostrows\end{tabular}
\endgroup
\end{table}

\begin{table*}[!t]
\centering
\caption{Contributions of backbone learning, evidence refinement and DDI feedback.}
\label{tab:refinement}
\evitable
\begin{tabular}{lG{7pt}rrrG{9pt}rrrG{9pt}rrr}
\toprule
& \multicolumn{4}{c}{MIMIC-IV} & \multicolumn{4}{c}{NWICU} & \multicolumn{4}{c}{SICdb} \\
\cmidrule(lr){2-5}\cmidrule(lr){6-9}\cmidrule(lr){10-13}
Variant & AP $\uparrow$ & F1 $\uparrow$ & Jaccard $\uparrow$ & DDI $\downarrow$ & AP $\uparrow$ & F1 $\uparrow$ & Jaccard $\uparrow$ & DDI $\downarrow$ & AP $\uparrow$ & F1 $\uparrow$ & Jaccard $\uparrow$ & DDI $\downarrow$ \\
\midrule
Backbone (20 epochs) & \evimeansd{0.7534}{0.0003} & \evimeansd{0.6337}{0.0015} & \evimeansd{0.4851}{0.0016} & \evimeansd{0.0611}{0.0012} & \evimeansd{0.6740}{0.0017} & \evimeansd{0.5913}{0.0003} & \evimeansd{0.4399}{0.0006} & \evimeansd{0.1166}{0.0036} & \evimeansd{0.7435}{0.0003} & \evimeansd{0.6450}{0.0011} & \evimeansd{0.4980}{0.0013} & \evimeansd{0.1295}{0.0012} \\
Backbone + 20 epochs & \evimeansd{0.7566}{0.0005} & \evimeansd{0.6383}{0.0005} & \evimeansd{0.4900}{0.0008} & \evimeansd{0.0622}{0.0021} & \evimeansd{0.6769}{0.0015} & \evimeansd{0.5932}{0.0023} & \evimeansd{0.4421}{0.0026} & \evimeansd{0.1166}{0.0033} & \evimeansd{0.7452}{0.0011} & \evimeansd{0.6440}{0.0019} & \evimeansd{0.4968}{0.0021} & \evimeansd{0.1296}{0.0017} \\
+ evidence refinement & \evimeansd{0.7774}{0.0005} & \evimeansd{0.6482}{0.0025} & \evimeansd{0.5015}{0.0029} & \evimeansd{0.0595}{0.0007} & \evimeansd{0.6905}{0.0002} & \evimeansd{0.6053}{0.0011} & \evimeansd{0.4545}{0.0014} & \evimeansd{0.1170}{0.0029} & \evimeansd{0.7507}{0.0007} & \evimeansd{0.6478}{0.0018} & \evimeansd{0.5009}{0.0021} & \evimeansd{0.1289}{0.0041} \\
+ prediction-only training & \evimeansd{0.7769}{0.0007} & \evimeansd{0.6459}{0.0021} & \evimeansd{0.4995}{0.0027} & \evimeansd{0.0574}{0.0004} & \evimeansd{0.6916}{0.0004} & \evimeansd{0.6028}{0.0012} & \evimeansd{0.4513}{0.0013} & \evimeansd{0.1127}{0.0015} & \evimeansd{0.7513}{0.0002} & \evimeansd{0.6451}{0.0019} & \evimeansd{0.4978}{0.0021} & \evimeansd{0.1260}{0.0015} \\
\evihighlight
\textbf{EviDrug: DDI feedback} & \evimeansd{0.7763}{0.0004} & \evimeansd{0.6452}{0.0022} & \evimeansd{0.4988}{0.0026} & \evimeansd{0.0522}{0.0014} & \evimeansd{0.6909}{0.0007} & \evimeansd{0.6026}{0.0016} & \evimeansd{0.4515}{0.0016} & \evimeansd{0.1043}{0.0032} & \evimeansd{0.7505}{0.0010} & \evimeansd{0.6425}{0.0023} & \evimeansd{0.4950}{0.0026} & \evimeansd{0.1181}{0.0052} \\
\bottomrule
\end{tabular}

\par\medskip
\begingroup\normalfont\normalsize
\makeatletter\def\@captype{figure}\makeatother
\includegraphics[width=\textwidth]{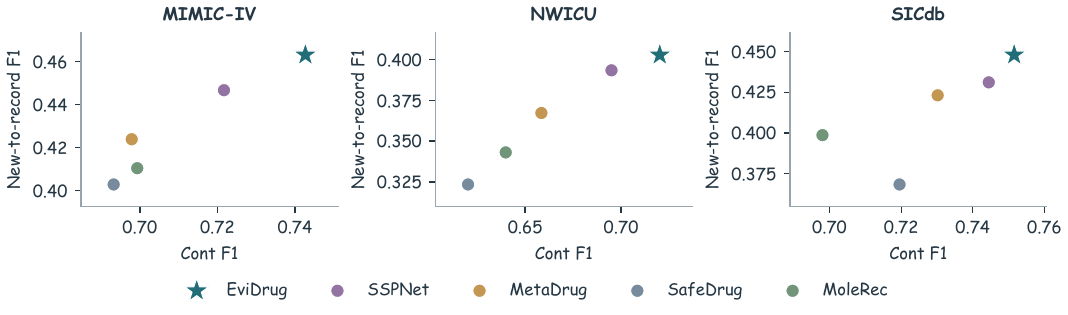}
\caption{Continued versus new-to-record medication recommendation (micro-F1).}
\label{fig:actions}
\endgroup
\end{table*}

\subsection{Compact Modeling and Batched Inference}
\label{sec:efficiency-results}
Table~\ref{tab:efficiency} evaluates the complete path from reading reference data to medication selection. EviDrug contains 7.766M, 2.142M, 1.069M parameters on MIMIC-IV, NWICU and SICdb, with cohort-specific vocabularies. Structured embedding and pooling concentrate clinical computation around visits. Refinement trains a shared 0.113M-parameter head that operates medication by medication.

Each cohort uses FP32, 128 fixed validation patients, $B=16$, one CPU thread and the median of three measured passes after warm-up. Method order varies between passes. Retrieval, features, collation, prediction and selection are included; startup is excluded. EviDrug reuses fixed training-case representations and association tables while retrieving cases and preparing query evidence online. Baselines retain native adaptation and decoding. Costs are amortized over batches on the shared host.
Compared with MetaDrug, EviDrug reduces parameter counts by \textbf{33.8\%, 49.6\%, 62.9\%} and CPU inference times by \textbf{51.7\%, 52.5\%, 32.7\%}, respectively, on MIMIC-IV, NWICU and SICdb.
These costs accompany the higher recommendation quality in Table~\ref{tab:published-baselines}. Dynamic interaction feedback contributes zero additional inference parameters.

The architecture allocates computation differently to clinical encoding and prescription interaction. Pooling creates one state per encounter, so clinical-code processing does not require self-attention over the full longitudinal token sequence. The medication-set block then models candidate interactions once. The refinement head reuses the resulting states and medication identities, adding a shared correction rather than a second patient encoder. This organization connects the reduced parameter counts to the higher AP in Table~\ref{tab:published-baselines}. The timing measurements include the online work needed to obtain case and association evidence, so the reported cost covers the complete prediction path used by the model.

\begin{table*}[!t]
\centering
\caption{Major-component ablations before (A) and after (B) DDI-feedback optimization.}
\label{tab:components}
\evitable
\begin{tabular}{lG{7pt}rrrG{9pt}rrrG{9pt}rrr}
\toprule
& \multicolumn{4}{c}{MIMIC-IV} & \multicolumn{4}{c}{NWICU} & \multicolumn{4}{c}{SICdb} \\
\cmidrule(lr){2-5}\cmidrule(lr){6-9}\cmidrule(lr){10-13}
Variant & AP $\uparrow$ & F1 $\uparrow$ & Jaccard $\uparrow$ & DDI $\downarrow$ & AP $\uparrow$ & F1 $\uparrow$ & Jaccard $\uparrow$ & DDI $\downarrow$ & AP $\uparrow$ & F1 $\uparrow$ & Jaccard $\uparrow$ & DDI $\downarrow$ \\
\midrule
\multicolumn{13}{@{}l}{\textit{A. After evidence refinement}} \\
\evihighlight
Full evidence & \evimeansd{0.7774}{0.0005} & \evimeansd{0.6482}{0.0025} & \evimeansd{0.5015}{0.0029} & \evimeansd{0.0595}{0.0007} & \evimeansd{0.6905}{0.0002} & \evimeansd{0.6053}{0.0011} & \evimeansd{0.4545}{0.0014} & \evimeansd{0.1170}{0.0029} & \evimeansd{0.7507}{0.0007} & \evimeansd{0.6478}{0.0018} & \evimeansd{0.5009}{0.0021} & \evimeansd{0.1289}{0.0041} \\
w/o longitudinal history & \evimeansd{0.7322}{0.0003} & \evimeansd{0.6054}{0.0026} & \evimeansd{0.4549}{0.0027} & \evimeansd{0.0608}{0.0006} & \evimeansd{0.5946}{0.0004} & \evimeansd{0.5099}{0.0019} & \evimeansd{0.3580}{0.0016} & \evimeansd{0.1348}{0.0042} & \evimeansd{0.7192}{0.0003} & \evimeansd{0.6106}{0.0006} & \evimeansd{0.4626}{0.0012} & \evimeansd{0.1414}{0.0029} \\
w/o training-case memory & \evimeansd{0.7656}{0.0016} & \evimeansd{0.6397}{0.0021} & \evimeansd{0.4917}{0.0022} & \evimeansd{0.0591}{0.0001} & \evimeansd{0.6888}{0.0009} & \evimeansd{0.6007}{0.0003} & \evimeansd{0.4494}{0.0006} & \evimeansd{0.1155}{0.0016} & \evimeansd{0.7504}{0.0026} & \evimeansd{0.6465}{0.0026} & \evimeansd{0.5005}{0.0029} & \evimeansd{0.1352}{0.0027} \\
w/o medication-set block & \evimeansd{0.7752}{0.0005} & \evimeansd{0.6462}{0.0015} & \evimeansd{0.4995}{0.0013} & \evimeansd{0.0582}{0.0006} & \evimeansd{0.6906}{0.0009} & \evimeansd{0.6037}{0.0007} & \evimeansd{0.4528}{0.0008} & \evimeansd{0.1154}{0.0024} & \evimeansd{0.7499}{0.0021} & \evimeansd{0.6485}{0.0022} & \evimeansd{0.5017}{0.0024} & \evimeansd{0.1250}{0.0016} \\
w/o clinical associations & \evimeansd{0.7545}{0.0013} & \evimeansd{0.6332}{0.0014} & \evimeansd{0.4855}{0.0016} & \evimeansd{0.0586}{0.0006} & \evimeansd{0.6756}{0.0019} & \evimeansd{0.5924}{0.0004} & \evimeansd{0.4413}{0.0007} & \evimeansd{0.1180}{0.0015} & \evimeansd{0.7444}{0.0006} & \evimeansd{0.6456}{0.0009} & \evimeansd{0.4986}{0.0012} & \evimeansd{0.1310}{0.0015} \\
\midrule
\multicolumn{13}{@{}l}{\textit{B. After DDI-feedback optimization}} \\
\evihighlight
Full evidence & \evimeansd{0.7763}{0.0004} & \evimeansd{0.6452}{0.0022} & \evimeansd{0.4988}{0.0026} & \evimeansd{0.0522}{0.0014} & \evimeansd{0.6909}{0.0007} & \evimeansd{0.6026}{0.0016} & \evimeansd{0.4515}{0.0016} & \evimeansd{0.1043}{0.0032} & \evimeansd{0.7505}{0.0010} & \evimeansd{0.6425}{0.0023} & \evimeansd{0.4950}{0.0026} & \evimeansd{0.1181}{0.0052} \\
w/o longitudinal history & \evimeansd{0.7306}{0.0008} & \evimeansd{0.6030}{0.0032} & \evimeansd{0.4531}{0.0037} & \evimeansd{0.0484}{0.0014} & \evimeansd{0.5939}{0.0006} & \evimeansd{0.5079}{0.0004} & \evimeansd{0.3562}{0.0002} & \evimeansd{0.1173}{0.0043} & \evimeansd{0.7197}{0.0005} & \evimeansd{0.6095}{0.0015} & \evimeansd{0.4617}{0.0019} & \evimeansd{0.1183}{0.0045} \\
w/o training-case memory & \evimeansd{0.7657}{0.0020} & \evimeansd{0.6369}{0.0015} & \evimeansd{0.4897}{0.0015} & \evimeansd{0.0516}{0.0012} & \evimeansd{0.6882}{0.0012} & \evimeansd{0.5964}{0.0022} & \evimeansd{0.4451}{0.0022} & \evimeansd{0.0963}{0.0084} & \evimeansd{0.7507}{0.0021} & \evimeansd{0.6437}{0.0015} & \evimeansd{0.4972}{0.0016} & \evimeansd{0.1215}{0.0015} \\
w/o medication-set block & \evimeansd{0.7747}{0.0005} & \evimeansd{0.6440}{0.0015} & \evimeansd{0.4973}{0.0014} & \evimeansd{0.0521}{0.0004} & \evimeansd{0.6906}{0.0008} & \evimeansd{0.6009}{0.0004} & \evimeansd{0.4497}{0.0007} & \evimeansd{0.1018}{0.0019} & \evimeansd{0.7491}{0.0016} & \evimeansd{0.6430}{0.0022} & \evimeansd{0.4958}{0.0024} & \evimeansd{0.1156}{0.0023} \\
w/o clinical associations & \evimeansd{0.7541}{0.0017} & \evimeansd{0.6288}{0.0020} & \evimeansd{0.4815}{0.0020} & \evimeansd{0.0522}{0.0012} & \evimeansd{0.6756}{0.0022} & \evimeansd{0.5876}{0.0022} & \evimeansd{0.4364}{0.0023} & \evimeansd{0.0953}{0.0067} & \evimeansd{0.7443}{0.0002} & \evimeansd{0.6437}{0.0008} & \evimeansd{0.4962}{0.0011} & \evimeansd{0.1246}{0.0033} \\
\bottomrule
\end{tabular}

\par\medskip
\begingroup\normalfont\normalsize
\makeatletter\def\@captype{figure}\makeatother
\includegraphics[width=\textwidth]{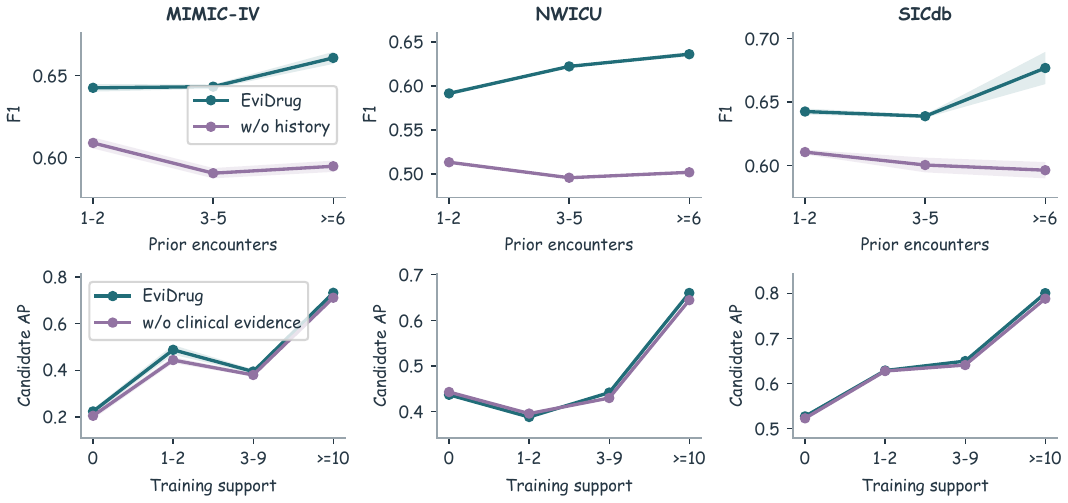}
\caption{Effects of history depth (top) and association support (bottom). Shading denotes $\pm$ one SD.}
\label{fig:mechanisms}
\endgroup
\end{table*}

\subsection{Contributions of the Backbone and Evidence Refinement}
Table~\ref{tab:refinement} distinguishes the contribution of evidence refinement from additional backbone training. It compares the clinical-memory backbone, the backbone trained for 20 more epochs, and evidence refinement with the backbone fixed. Prediction-only training and DDI-feedback optimization start from the same refinement model within each seed, isolating the contribution of the interaction objective.

\begin{table*}[!t]
\centering
\caption{Fixed versus dynamic DDI regularization with matched training budgets.}
\label{tab:weights}
\evitable
\begin{tabular}{lG{7pt}rrrG{9pt}rrrG{9pt}rrr}
\toprule
& \multicolumn{4}{c}{MIMIC-IV} & \multicolumn{4}{c}{NWICU} & \multicolumn{4}{c}{SICdb} \\
\cmidrule(lr){2-5}\cmidrule(lr){6-9}\cmidrule(lr){10-13}
Variant & AP $\uparrow$ & F1 $\uparrow$ & Jaccard $\uparrow$ & DDI $\downarrow$ & AP $\uparrow$ & F1 $\uparrow$ & Jaccard $\uparrow$ & DDI $\downarrow$ & AP $\uparrow$ & F1 $\uparrow$ & Jaccard $\uparrow$ & DDI $\downarrow$ \\
\midrule
Fixed $\lambda=0$ & \evimeansd{0.7769}{0.0007} & \evimeansd{0.6459}{0.0021} & \evimeansd{0.4995}{0.0027} & \evimeansd{0.0574}{0.0004} & \evimeansd{0.6916}{0.0004} & \evimeansd{0.6028}{0.0012} & \evimeansd{0.4513}{0.0013} & \evimeansd{0.1127}{0.0015} & \evimeansd{0.7513}{0.0002} & \evimeansd{0.6451}{0.0019} & \evimeansd{0.4978}{0.0021} & \evimeansd{0.1260}{0.0015} \\
Fixed $\lambda=0.05$ & \evimeansd{0.7763}{0.0002} & \evimeansd{0.6449}{0.0029} & \evimeansd{0.4985}{0.0028} & \evimeansd{0.0528}{0.0012} & \evimeansd{0.6915}{0.0009} & \evimeansd{0.6017}{0.0018} & \evimeansd{0.4505}{0.0019} & \evimeansd{0.1010}{0.0025} & \evimeansd{0.7511}{0.0009} & \evimeansd{0.6426}{0.0025} & \evimeansd{0.4953}{0.0027} & \evimeansd{0.1202}{0.0023} \\
Fixed $\lambda=0.1$ & \evimeansd{0.7759}{0.0005} & \evimeansd{0.6443}{0.0026} & \evimeansd{0.4975}{0.0028} & \evimeansd{0.0491}{0.0005} & \evimeansd{0.6901}{0.0003} & \evimeansd{0.6018}{0.0016} & \evimeansd{0.4507}{0.0018} & \evimeansd{0.1028}{0.0013} & \evimeansd{0.7506}{0.0003} & \evimeansd{0.6427}{0.0027} & \evimeansd{0.4954}{0.0030} & \evimeansd{0.1179}{0.0024} \\
Fixed $\lambda=1$ & \evimeansd{0.7774}{0.0005} & \evimeansd{0.6482}{0.0025} & \evimeansd{0.5015}{0.0029} & \evimeansd{0.0595}{0.0007} & \evimeansd{0.6905}{0.0002} & \evimeansd{0.6053}{0.0011} & \evimeansd{0.4545}{0.0014} & \evimeansd{0.1170}{0.0029} & \evimeansd{0.7507}{0.0007} & \evimeansd{0.6478}{0.0018} & \evimeansd{0.5009}{0.0021} & \evimeansd{0.1289}{0.0041} \\
\evihighlight
\textbf{Dynamic $\lambda$ (EviDrug)} & \evimeansd{0.7763}{0.0004} & \evimeansd{0.6452}{0.0022} & \evimeansd{0.4988}{0.0026} & \evimeansd{0.0522}{0.0014} & \evimeansd{0.6909}{0.0007} & \evimeansd{0.6026}{0.0016} & \evimeansd{0.4515}{0.0016} & \evimeansd{0.1043}{0.0032} & \evimeansd{0.7505}{0.0010} & \evimeansd{0.6425}{0.0023} & \evimeansd{0.4950}{0.0026} & \evimeansd{0.1181}{0.0052} \\
\bottomrule
\end{tabular}

\par\medskip
\begingroup\normalfont\normalsize
\makeatletter\def\@captype{figure}\makeatother
\includegraphics[width=\textwidth]{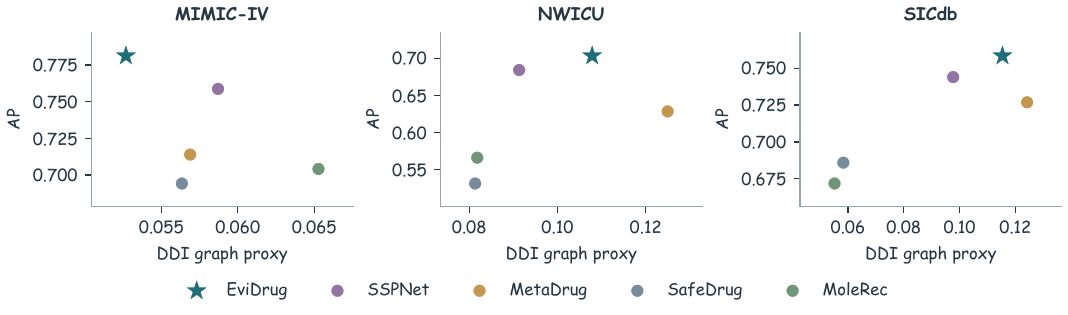}
\caption{AP--DDI trade-off for patients with at least 11 prior medications.}
\label{fig:burden}
\endgroup
\end{table*}

Evidence refinement increases AP over the original backbone by 0.0240, 0.0165, 0.0072 on MIMIC-IV, NWICU and SICdb, respectively. Its AP advantage over extended backbone training is 0.0208, 0.0136, 0.0056. These comparisons support evidence refinement \textbf{beyond additional backbone training}. DDI-feedback optimization then regulates interactions in the refined predictions.

The extended-backbone control separates training duration from the use of additional evidence. It gives the original predictor the same additional epoch budget as the refinement stage, while refinement keeps that predictor fixed and learns how to revise its medication scores. The larger AP increase from refinement indicates that the gain comes from the evidence update. Zero-initialization makes the comparison especially direct: refinement starts with exactly the backbone's logits, then changes them through the learned correction. The final feedback stage tests a separate role, namely adapting interactions after the improved prescription has been established.
Complete-module ablations identify the evidence sources that drive these gains. History removal uses the current encounter and state-only retrieval; memory removal deletes case retrieval and donor encoding; set-block removal deletes medication-set attention. Each of these variants learns its own backbone and refinement stages. Clinical-association deletion shares the full backbone and refits the affected head. All variants use three seeds with the same stage budgets and selection rules.
Table~\ref{tab:components} reports component contributions after evidence refinement (panel A) and after DDI-feedback optimization (panel B). The two views distinguish the predictive contribution of each component from its behavior under interaction regularization.

\textbf{Longitudinal history and aggregate clinical associations} account for the largest AP contributions. History removal reduces AP by 0.0457, 0.0970, 0.0308 on MIMIC-IV, NWICU and SICdb; removing clinical associations reduces it by 0.0222, 0.0153, 0.0062. Case memory contributes an additional 0.0106 AP on MIMIC-IV. The complete matrix shows how case evidence and set interactions complement these main clinical signals under the shared training objective.

History and clinical associations provide different references for the same candidate. Earlier prescriptions describe an individual patient's recorded treatment trajectory, while diagnosis--medication counts summarize evidence across training encounters. The substantial history effect on NWICU and the association effect on MIMIC-IV demonstrate that their relative value varies across cohorts. The full model makes both available to a medication-specific gate. Panels A and B also distinguish predictive contributions from changes induced by interaction regularization: the same evidence source is assessed before and after feedback. Each removal is interpreted against its corresponding full model, since the effects of jointly trained components need not add linearly.

\begin{table*}[!t]
\centering
\caption{Continued and new-to-record medication recommendation (micro-F1).}
\label{tab:clinical-actions}
\evitable
\begin{tabular}{lG{8pt}rrG{14pt}rrG{14pt}rr}
\toprule
& \multicolumn{3}{c}{MIMIC-IV} & \multicolumn{3}{c}{NWICU} & \multicolumn{3}{c}{SICdb} \\
\cmidrule(lr){2-4}\cmidrule(lr){5-7}\cmidrule(lr){8-10}
Method & Cont & New & Avg & Cont & New & Avg & Cont & New & Avg \\
\midrule
RETAIN & 0.6999 & 0.4218 & 0.5609 & 0.6276 & 0.3638 & 0.4957 & 0.7271 & 0.4258 & 0.5765 \\
Deepr & 0.6742 & 0.3849 & 0.5296 & 0.6178 & 0.3478 & 0.4828 & 0.7289 & 0.4214 & 0.5752 \\
Transformer & 0.6718 & 0.3451 & 0.5085 & 0.6293 & 0.3503 & 0.4898 & 0.7254 & 0.4198 & 0.5726 \\
GAMENet & 0.6955 & 0.3858 & 0.5406 & 0.6572 & 0.3411 & 0.4992 & 0.7282 & 0.4073 & 0.5677 \\
MeLU & 0.6490 & 0.3560 & 0.5025 & 0.6083 & 0.3271 & 0.4677 & 0.7157 & 0.3937 & 0.5547 \\
MICRON & 0.6952 & 0.4281 & 0.5617 & 0.6199 & 0.3472 & 0.4835 & 0.7312 & 0.4337 & 0.5825 \\
SafeDrug & 0.6932 & 0.4029 & 0.5481 & 0.6202 & 0.3234 & 0.4718 & 0.7195 & 0.3684 & 0.5440 \\
COGNet & 0.6108 & 0.3669 & 0.4888 & 0.6655 & 0.2780 & 0.4718 & 0.7395 & 0.3647 & 0.5521 \\
MoleRec & 0.6993 & 0.4104 & 0.5549 & 0.6399 & 0.3431 & 0.4915 & 0.6980 & 0.3987 & 0.5483 \\
ARCI & 0.6581 & 0.3062 & 0.4821 & 0.6817 & 0.3198 & 0.5008 & 0.7408 & 0.3909 & 0.5659 \\
SSPNet & 0.7217 & 0.4466 & 0.5842 & 0.6950 & 0.3934 & 0.5442 & 0.7444 & 0.4310 & 0.5877 \\
MetaDrug & 0.6979 & 0.4239 & 0.5609 & 0.6585 & 0.3672 & 0.5129 & 0.7302 & 0.4231 & 0.5766 \\
\evihighlight
\textbf{EviDrug} & \evimeansd{\textbf{0.7427}}{0.0024} & \evimeansd{\textbf{0.4630}}{0.0022} & \evimeansd{\textbf{0.6028}}{0.0006} & \evimeansd{\textbf{0.7202}}{0.0003} & \evimeansd{\textbf{0.4029}}{0.0040} & \evimeansd{\textbf{0.5615}}{0.0022} & \evimeansd{\textbf{0.7515}}{0.0013} & \evimeansd{\textbf{0.4479}}{0.0042} & \evimeansd{\textbf{0.5997}}{0.0027} \\
\bottomrule
\end{tabular}
\end{table*}

\subsection{DDI-Feedback Optimization with Preserved Predictive Quality}
Table~\ref{tab:weights} evaluates DDI-feedback optimization with matched initialization and training budgets. Fixed coefficients $0/0.05/0.1/1$ and the feedback rule use the same seed-specific refinement model, fresh optimizer, 20-epoch budget, and prediction-quality and prescription-size criteria. The dynamic coefficient starts at 1 and adapts from training queries.

Under the shared validation criteria, fixed $\lambda=1$ selects the initial refinement model in all nine runs. DDI feedback adapts the penalty to the predicted medication combinations and selects a model updated by feedback optimization in every run. This comparison supports adapting regularization strength to the predictions produced during training.
Compared with matched prediction-only training, DDI-feedback optimization reduces mean test DDI by \textbf{9.0\%, 7.5\%, 6.2\%} on MIMIC-IV, NWICU and SICdb, while the absolute AP and F1 changes stay \textbf{below 0.003 in every cohort}. The fixed-weight controls characterize the prediction--interaction trade-off. A common feedback rule achieves this reduction across cohorts while retaining the predictive quality of evidence refinement.

These controls distinguish adaptation of the objective from the effect of further optimization. Prediction-only training receives the same extra updates, but its interaction coefficient remains zero. Fixed-weight runs apply the interaction penalty throughout the same budget, while feedback changes its strength according to the combinations produced by the current head. The prediction loss, backbone and validation criteria are shared. Consequently, the comparison evaluates how interaction control is allocated during learning. The dynamic model combines this training-time adjustment with the same inference path as the initial refinement model; the selected coefficient does not become an additional input or computation at recommendation time.

\subsection{Evidence for Continued and New-to-Record Medications}
The clinical analyses ask where the prediction gains arise. We first distinguish target ATC3 classes seen in earlier eligible encounters (Cont) from classes absent from the available history (New). Table~\ref{tab:clinical-actions} evaluates both candidate sets with micro-F1 and their unweighted average (Avg), testing whether EviDrug supports treatment persistence and medications beyond the recorded history.

EviDrug leads all twelve comparators in \textbf{Cont, New and Avg F1 on every cohort}. Against SSPNet, its average gains are 0.0187, 0.0173, 0.0120. Figure~\ref{fig:actions} shows that the improvement spans both strata; positive supports and error counts are retained in the artifacts.

The two strata make the treatment-history contribution more specific. A continued class has appeared in the available record, so its prediction can draw on evidence of previous use. A new-to-record class must be supported without that positive history indicator. EviDrug improves both: its Cont F1 is 0.7427/0.7202/0.7515 and its New F1 is 0.4630/0.4029/0.4479 in cohort order. This pattern supports combining history with current clinical and case evidence. New-to-record refers to the observed patient history, not to a newly introduced drug or an unseen medication class. The separate strata therefore characterize how the model reconstructs treatment combinations from the information available at the query encounter.

\paragraph{History depth and association support}
Figure~\ref{fig:mechanisms} links the benefit of longitudinal modeling to the amount of recorded history. Patients are grouped by 1--2, 3--5 and at least 6 prior encounters. F1 gaps between the full model and history removal grow from 0.0338/0.0783/0.0320 in the shortest-history group to 0.0664/0.1342/0.0806, in the longest, in cohort order. This pattern supports the encoder in extracting useful context from longer treatment trajectories.
The bottom panels test aggregate clinical evidence. Support is the maximum training diagnosis--medication co-occurrence count for each candidate, binned as 0, 1--2, 3--9 and at least 10. Within identical candidate strata, clinical-association refinement improves pooled AP by 0.0211/0.0157/0.0122 at support at least 10. The contribution depends on support, complementing individual-case evidence with population-level recording patterns. Candidate AP pools patient--medication pairs within each bin.

The two panels examine complementary forms of evidence availability. Longer records provide more clinical states and prescriptions for history attention, while larger co-occurrence counts provide a stronger empirical reference for the association branch. Their gains arise through different paths in Figure~\ref{fig:architecture}: history changes the patient context and medication identity, whereas associations change the evidence supplied to refinement. Grouping patients and candidates by these quantities connects the overall improvements to the information the corresponding component can actually use. The comparison within each support bin also keeps the evaluated candidate set fixed when the association branch is removed.

\paragraph{Prior medication burden and interaction exposure}
The high-burden analysis tests prediction quality for patients with larger preceding prescriptions. Burden is defined from the previous encounter using bins 0--5, 6--10 and at least 11 medications. Figure~\ref{fig:burden} evaluates the last group, showing AP alongside the DDI operating point. Table~\ref{tab:burden-pairs} adds prescription size and interacting-pair counts, both averaged per patient and over three seeds, to characterize the complete predicted combination.

\begin{table}[!t]
\centering
\caption{EviDrug predictions in the high prior-burden group.}
\label{tab:burden-pairs}
\evitable
\begin{tabular}{@{}lrrrr@{}}
\toprule
Dataset & Patients & Medications & DDI & Pairs \\
\midrule
MIMIC-IV & 13,608 & 27.8010 & 0.0527 & 29.3483 \\
NWICU & 1,307 & 15.6993 & 0.1079 & 13.9666 \\
SICdb & 322 & 13.9917 & 0.1154 & 10.8468 \\
\bottomrule
\end{tabular}
\end{table}

EviDrug maintains the highest AP among the displayed methods in the high-burden group of every cohort. Its gains over SSPNet are 2.25, 1.91, 1.44 percentage points on MIMIC-IV, NWICU and SICdb, extending the overall ranking advantage to these larger-prescription histories.

Prior burden is determined before the query, so membership in this analysis does not depend on how many medications a model predicts. Within the same patient group, Figure~\ref{fig:burden} places ranking quality beside the interaction rate, and Table~\ref{tab:burden-pairs} records the size and pair exposure of EviDrug's complete combination. The quantities serve different purposes: a prescription with $m$ medications contains $m(m-1)/2$ unordered pairs, and DDI measures the interacting fraction of those pairs. Reporting counts together with the rate connects the interaction statistic to the size of the predicted prescription. This is particularly relevant when the task involves many concurrent medication classes.

\paragraph{Prediction gains at equal prescription size}
Top-$k$ evaluation tests whether recommendation gains persist when \textbf{prescription size is held equal}. Values of $k$ come from rounded training-prescription-size quantiles 0.25/0.5/0.75/0.9. Averaged across these values, EviDrug improves F1 over SSPNet by 0.0092/0.0132/0.0155. The gains therefore extend beyond the fixed-threshold operating point. Interaction rates at each $k$ and threshold curves for 0.1--0.9 are supplied with the complete results.

Holding $k$ equal makes the comparison depend on which medications occupy the selected positions. A higher F1 in this setting reflects better alignment with the recorded combination at the same recommendation budget. Together with the fixed-threshold results, the analysis links EviDrug's AP advantage to useful changes in the selected candidates across several prescription sizes. The $k$ values come from training data and are shared by methods within a cohort.
\paragraph{Consistency across paired test patients}
A patient-paired bootstrap with 2,000 resamples assesses the AP advantage over SSPNet. Each resample compares all three EviDrug seeds with the same fixed baseline checkpoint and averages the differences. The resulting percentile intervals describe patient-sampling variability conditional on these checkpoints.
The AP gain and its 95\% patient-bootstrap interval are: MIMIC-IV +0.0226 [+0.0212, +0.0242]; NWICU +0.0206 [+0.0161, +0.0251]; SICdb +0.0171 [+0.0114, +0.0227]. \textbf{Every interval lies above zero}, supporting a consistent AP advantage across the evaluated test patients.

The paired analysis retains each patient's contribution to the comparison. A resample containing more high-burden patients changes both models' evaluated case mix together. Its positive lower endpoint in every cohort supports the consistency in Table~\ref{tab:published-baselines}. Together with the fixed-size evaluation, the result connects the improvement to both the patient population and the choice of medication candidates at a common prescription size. The two analyses test complementary aspects of the aggregate advantage.

\FloatBarrier
\par\medskip\noindent\begin{minipage}{\columnwidth}
\makeatletter\def\@captype{figure}\makeatother
\centering
\includegraphics[width=\columnwidth]{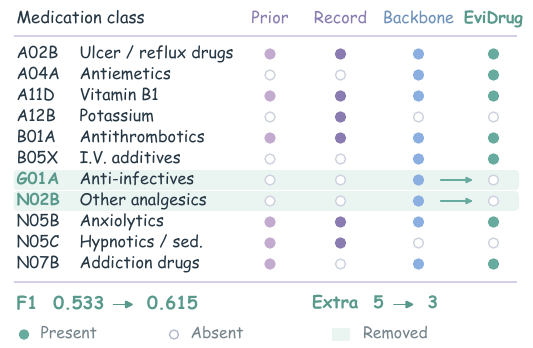}
\caption{Prescription refinement for one NWICU encounter. Shading marks removed extra predictions.}
\label{fig:case-example}
\end{minipage}\par\medskip

\paragraph{Prescription refinement in one encounter}
Figure~\ref{fig:case-example} shows an anonymized NWICU encounter nearest the median patient F1 within the 6--10 prior-medication stratum (seed 0). EviDrug preserves four correctly predicted classes and removes two extra predictions relative to the complete backbone, increasing patient \textbf{F1 from 0.533 to 0.615}.

The medication rows show where this correction occurs. The recorded ulcer/reflux, vitamin B1, antithrombotic and anxiolytic classes remain selected. Two additional backbone predictions, G01A and N02B, cross below the common threshold after refinement. Thus the improvement in this encounter comes from more selective use of the evidence while retaining the four matched classes. The figure displays every class present in the history, recorded target or either prediction, so the set-level comparison includes the full pattern of agreements and changes.

Refinement changes individual scores in different directions: A11D rises from 0.533 to 0.775, G01A falls from 0.411 to 0.240, and B01A stays near 0.761. This behavior connects the example to the evidence gate in Figure~\ref{fig:architecture}. Shared patient context is combined with medication-specific support, producing distinct updates within the same encounter. At the common threshold, these corrections preserve matched classes and remove extra selections.

\section{Conclusion}
EviDrug combines a clinical-memory backbone, bounded medication-specific refinement and DDI-feedback optimization. Patient history and aggregate clinical associations provide complementary evidence, with stronger history benefits in longer records. Structured pooling, one medication-set interaction block and a shared refinement head form a compact recommendation path; dynamic feedback adds no inference parameters. Across three complete eligible cohorts, EviDrug leads twelve baselines in AP, F1 and Jaccard. Matched controls show reduced DDI with preserved predictive quality. These gains are achieved with fewer parameters and lower inference cost than MetaDrug across all three cohorts.

\section*{AI Assistance Disclosure}
DeepSeek V4.1-Flash, GLM-5.3-Flash and GPT were used for translation and language polishing. The authors are responsible for the content, references and reported results.
\IEEEtriggeratref{15}
\bibliographystyle{IEEEtran}
\bibliography{references}
\end{document}